\documentclass[11pt]{article}

\usepackage[preprint]{acl}

\usepackage{times}
\usepackage{latexsym}

\usepackage[T1]{fontenc}

\usepackage[utf8]{inputenc}

\usepackage{microtype}

\usepackage{inconsolata}

\usepackage{graphicx}
\usepackage{amsmath}
\usepackage{amssymb}
\usepackage{mathtools}
\usepackage{amsthm}
\usepackage{hyperref}
\usepackage{url}
\usepackage{xspace,mfirstuc,tabulary}
\usepackage{booktabs}  
\usepackage{longtable} 
\usepackage{array}     
\usepackage{subcaption}  
\usepackage{makecell}    
\usepackage{xspace}
\usepackage{wrapfig}
\usepackage{multicol}
\usepackage{multirow}
\usepackage{tikz}
\usepackage{caption}
\usepackage{xcolor} 
\usepackage{tabularx} 
\usepackage{makecell} 

\DeclareRobustCommand{\coloredlines}{%
    \tikz[baseline=0.ex]\draw[blue, thick] (0,0) -- (0.6em,0); 
    and  
    \tikz[baseline=0.ex]\draw[red, thick] (0,0) -- (0.6em,0); 
}

\newcommand{\comp}{\textsc{Comp}\xspace}
\newcommand{\para}{\textsc{Mem}\xspace}
\newcommand{\context}{\textsc{Ctx}\xspace}

\newcommand{\sftatomic}{$\textsc{SFT}_{\textsc{Mem+Ctx}}$\xspace}
\newcommand{\sftcomp}{$\textsc{SFT}_{\textsc{Comp}}$\xspace}
\newcommand{\rlcomp}{$\textsc{RL}_{\textsc{Comp}}$\xspace}
\newcommand{\rlatomic}{$\textsc{RL}_{\textsc{Mem+Ctx}}$\xspace}

\title{\textit{Atomic Skills are the Prerequisite}: When Reinforcement Learning Synthesizes Compositional Reasoning, and When It Only Amplifies}

\author{\textbf{Sitao Cheng}$^{1\dagger}$,~\textbf{Xunjian Yin}$^{2}$,~\textbf{Ruiwen Zhou}$^{3}$,~ 
    \textbf{Yuxuan Li}$^{1}$, \\
    \textbf{Xinyi Wang}$^{4}$,~\textbf{Liangming Pan}$^{5\dagger}$,~\textbf{William Yang Wang}$^{6}$,~ \textbf{Victor Zhong}$^{1\dagger}$\\[0.5em]
    $^1$University of Waterloo \quad $^2$Duke University \quad $^3$National University of Singapore \\
    $^4$Princeton University \quad $^5$Peking University \quad $^6$University of California, Santa Barbara\\[0.5em]
    \texttt{\{sitao.cheng, victor.zhong\}@uwaterloo.ca} \quad \texttt{liangmingpan@pku.edu.cn}
}

\begin{document}
\maketitle
\renewcommand{\thefootnote}{\fnsymbol{footnote}}
\footnotetext[2]{Corresponding author.}
\renewcommand{\thefootnote}{\arabic{footnote}} 

\begin{abstract}
Does Reinforcement Learning (RL) merely amplify existing skills, or synthesize novel skills? We investigate this question through the lens of \textit{Complementary Reasoning}: the critical practical capability of integrating internal knowledge with external context, a prerequisite for reliable Continual Learning and Retrieval-Augmented Generation.
To avoid pre-training contamination, we construct a controlled semantic-synthetic dataset of biographies and decompose this capability into two atomic skills: \textit{Parametric Reasoning} (retrieving facts encoded in model weights) and \textit{Contextual Reasoning} (processing novel in-context information).
We present two findings.
First, models supervised directly on the composite task reach high accuracy on seen facts and reasoning paths (90\%) but collapse on novel facts and reasoning paths (18\%), indicating that Supervised Fine-Tuning (SFT) relies on rote memorization rather than genuine skill integration. Second, RL bridges this generalization gap, acting as a skill \textit{synthesizer} rather than a mere amplifier---but only under a strict prerequisite: it synthesizes new composite strategies only when the base model has first mastered the independent atomic skills via SFT.
These results suggest that decoupled atomic training followed by RL offers a scalable path to complex novel reasoning. Code and data are available at \href{https://github.com/sitaocheng/from_atomic_to_composite}{link}.
\end{abstract}

\section{Introduction}

\begin{figure*}
    \setlength{\abovecaptionskip}{0.cm}
    \setlength{\belowcaptionskip}{-0.5cm}
    \centering
    \includegraphics[width=\linewidth]{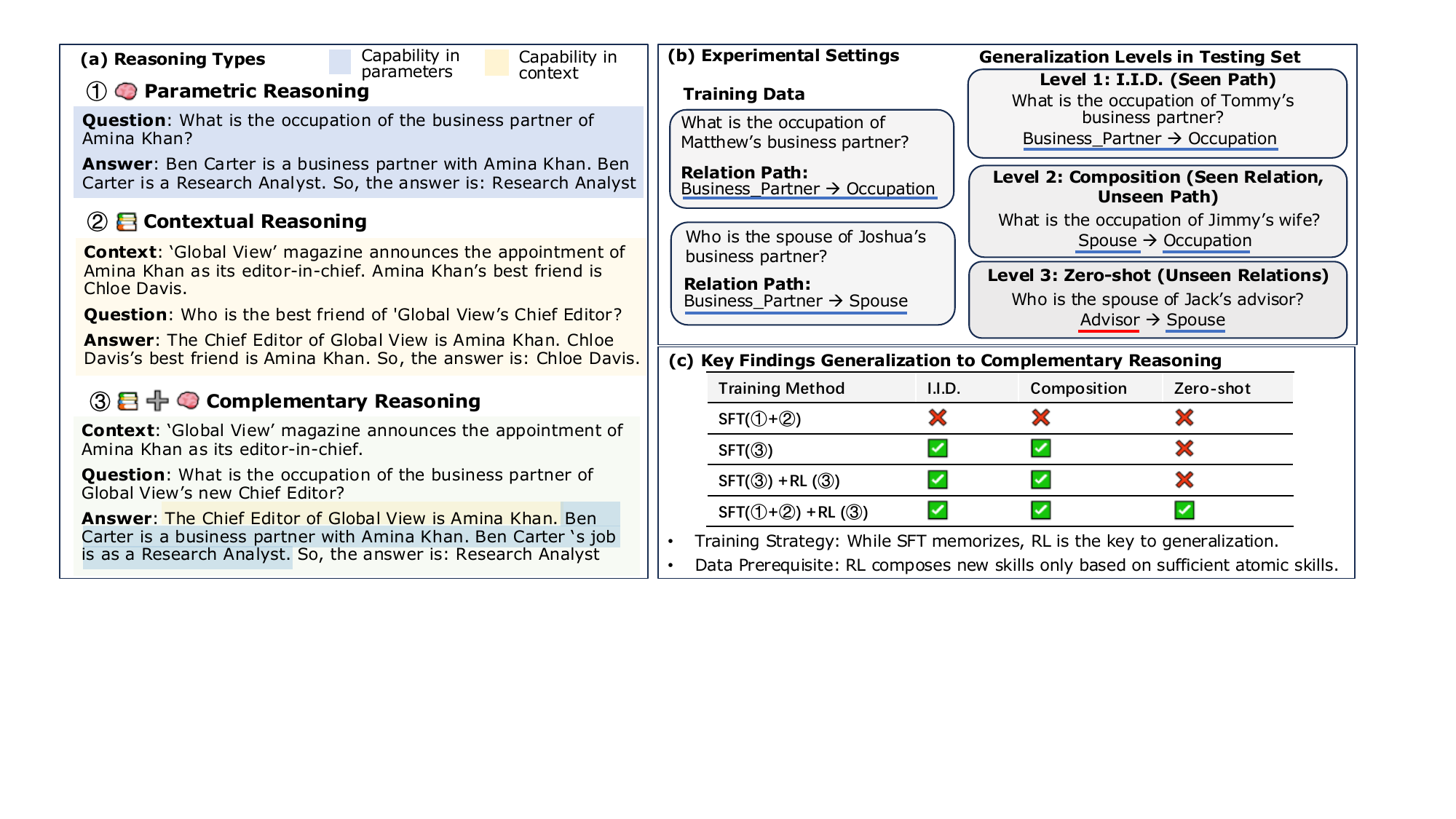}

    \caption[Table with colored lines]{Our settings and findings. (a) Examples of Complementary Reasoning requiring both Parametric and Contextual skills. (b) Evaluation protocol across three levels of difficulty. \protect\coloredlines denotes seen and unseen pattern, respectively. (c) The SFT Generalization Paradox: Models trained on atomic skills generalize better via RL than models trained directly on the composite task.}
    \label{fig:main_figure}
\vspace{-2pt}
\end{figure*}

The rapid evolution of Large Language Models (LLMs) is driven by Supervised Fine-Tuning (SFT) followed by Reinforcement Learning (RL) \citep{achiam2023gpt4,team2024gemini}. While SFT imparts foundational knowledge, its maximum likelihood objective inherently favors \textit{memorization}, often limiting out-of-distribution generalization~\citep{chu2025sft}. Conversely, RL is hypothesized to shift model distributions toward goal-oriented problem-solving \citep{schulman2017proximal,guo2025deepseekr1}.
Yet its fundamental mechanisms remain fiercely debated: some argue that RL synthesizes genuinely new reasoning skills \citep{yuan2025fxgxfgxllms,liu2025prorl}, while others characterize RL as an amplifier of probability mass already present in the post-SFT distribution \citep{wu2025invisible,yue2025does}.
This tension motivates two research questions: \textit{\textbf{1)} Does RL act as a synthesizer of new reasoning skills, or merely as an amplifier of existing ones? \textbf{2)} What training conditions are strictly necessary for RL-driven generalization to complex reasoning?}

Settling this debate requires a testbed that {\textit{a)}} is verifiably contamination-free, {\textit{b)}} clearly disentangles skills, and {\textit{c)}} admits genuinely novel relational compositions rather than mere extrapolation of seen patterns. Existing testbeds fall short on all three. Open-domain benchmarks (\textit{e.g.,} HotpotQA \citep{yang2018hotpotqa}) suffer from pre-training contamination, making it impossible to distinguish reasoning from memorized shortcuts. Controlled math and coding benchmarks  instead conflate reasoning mechanics with memory retrieval, making it hard to clearly attribute gains to either. Furthermore, ``new skills'' are often defined as the extrapolation of existing ones (\textit{e.g.,} longer reasoning chains) over a narrow set of operations (\textit{e.g.,} $+, -, *$) \citep{zhang2025interplay}, rather than the synthesis of truly unseen and meaningful relational patterns.

We propose \textbf{Complementary Reasoning (\comp)} as an ideal testbed satisfying all three criteria. \comp forces the model to bridge internal parametric knowledge with external contextual information to form novel logical connections, a core mechanism required for \textit{continual learning (CL)}. In Figure~\ref{fig:main_figure}a, answering \textit{``the occupation of the business partner of Global View's chief editor''} requires integrating external information (\textit{Amina Khan is the chief...}) with parametric knowledge (\textit{Ben Carter is a business...}). Unlike math, where rules (\textit{e.g.,} the distributive property) are fixed, this task demands the complementary composition of parametric and contextual knowledge simultaneously---a known challenge for LLMs~\citep{cheng2024understanding}. We accordingly decouple \comp into two atomic skills: \textbf{Parametric Reasoning ($\mathcal{C}_{\textrm{MEM}}$)}, relying on facts encoded in model weights, and \textbf{Contextual Reasoning ($\mathcal{C}_{\textrm{CTX}}$)}, relying on novel information provided in the context window. To remove contamination while preserving real-world generality, we avoid \textit{fully symbolic synthesis} (which strips away natural-language semantics) and instead employ \textbf{Semantic Synthesis} with Novel Entities \citep{allen2023physics3.1}: we retain standard real-world relations (\textit{e.g.,} spouse) but instantiate them with novel entities via synthetic biographies. This mimics the cognitive bottleneck of real-world CL and RAG systems---combining novel facts with pre-trained semantic understanding---while strictly enforcing the boundary between parametric and contextual knowledge. We evaluate across three generalization levels (Figure~\ref{fig:main_figure}b): \textit{IID} (seen paths), \textit{Composition} (unseen paths of seen relations), and \textit{Zero-shot} (unseen relations).

With controlled multi-hop QA pairs, we vary training strategies to identify the optimal settings for generalization (Figure~\ref{fig:main_figure}c). Our experiments uncover the \textit{SFT Generalization Paradox}: contrary to expectation, training directly on the composite target task via SFT fails to generalize out-of-distribution despite high in-distribution performance. However, RL resolves this paradox by acting as a powerful \textit{synthesizer} of reasoning skills---particularly in zero-shot where relation paths are entirely novel---\textit{provided the model first acquires sufficient atomic skills}. This challenges the view of RL as merely a probability amplifier, and suggests a scalable recipe: rather than collecting expensive complex reasoning traces, one can efficiently teach atomic skills via SFT and then leverage RL to unlock generalization for complex reasoning. We summarize our key contributions:

\noindent$\bullet$ We define \textbf{{Complementary Reasoning}} and introduce a semantic-synthetic dataset that decouples reasoning into atomic Parametric and Contextual skills, enabling contamination-free evaluation that mirrors real-world CL and RAG bottlenecks.

\noindent$\bullet$ We provide empirical evidence that while SFT drives memorization, RL is essential for generalization, specifically to the structural unseen zero-shot combinations that SFT fails to resolve.

\noindent$\bullet$ We uncover a prerequisite for RL generalization: RL synthesizes new complex skills \textbf{only when the base model possesses sufficient atomic skills}, vastly and efficiently outperforming direct SFT on composite data, regardless of data amount.

\section{Related Work}

\textbf{Roles of Post-training Strategies} 
LLMs capabilities are developed via SFT followed by RL.
While SFT establishes foundation knowledge, its maximum likelihood estimation often favors \textit{memorization} over generalization \citep{wang2025generalization,chu2025sft}. Conversely, RL uses reward signals to steer distributions toward goal-oriented problem-solving \citep{schulman2017proximal}. A central debate persists: does RL incentivize genuine skill acquisition \citep{yuan2025fxgxfgxllms,liu2025prorl}, or merely amplify pre-existing patterns \citep{wu2025invisible,yue2025does,yangdemystifying,setlur2025e3learningexploreenables}. One reason this tension remains unresolved is that prior works evaluate on math or code benchmarks with relatively narrow definition of generalization. Our work differs by isolating memory with reasoning using knowledge-intensive tasks. We find that RL fosters new skills under a prerequisite that the base model possesses sufficient atomic capabilities.

\noindent\textbf{Knowledge-intensive and Compositional Reasoning} Reasoning with knowledge—particularly multi-hop QA \citep{yang2018hotpotqa}—serves as a primary testbed for intelligence, requiring to retrieve facts and logically \textit{compose} them into a final answer \citep{jin2025disentangling}. While prior work focuses on inference-time by step-by-step prompting \citep{huang2023question,cheng2024call,gutierrez2025rag} or architectural interventions \citep{gu2023don,cheng2025differentiable}, they often obscure the underlying learning dynamics. A separate strand documents that LLMs frequently struggle when integrating novel facts with established parametric knowledge \citep{cheng2024understanding,yin2023alcuna}, but the underlying causes are often confounded by data contamination in open-domain benchmarks.
None of them expose the \textit{training-time prerequisites} for generalization. We differ by using complementary reasoning as a \textit{controlled mechanistic probe}, identifying the strict post-training conditions for compositional generalization to emerge.

\noindent\textbf{Behavioral Study with Synthetic Data} Synthetic data offers strict control over knowledge sufficiency, task complexity and contamination, becoming an important tool for analyzing learning dynamics in LLMs \citep{allen2023physics3.1,allen2023physics3.2}. Existing studies examine knowledge storage, extraction, or manipulation in isolation, relying on fully symbolic probes without natural-language semantics \citep{wang2024grokked}, which limits real-world transferability. We extend this with \textit{semantic synthesis with novel entities}: real-world relational semantics with novel factual instantiations. Moreover, our framework acts as a behavioral probe of post-training strategies under controlled splits (IID, Composition, Zero-shot), enabling rigorous evaluation of generalization to OOD relational patterns.

\section{Problem Definition}
\label{sec:problem_def}

To investigate the limitations of SFT and the mechanism of RL generalization, we first formalize the mechanical components of reasoning.
We define the atomic skills, the composite task, and the specific gradations of difficulty used to stress-test capabilities beyond the training distribution.

\textbf{Reasoning Types} We define \textbf{\textit{Complementary Reasoning}} (\textbf{\comp}) as a task requiring \textit{both} parametric \textit{and} contextual skills, a core mechanism for \textit{continual learning}.
Accordingly, we have \textbf{\textit{Parametric Reasoning} (\para)} requiring skills encoded in model weights and \textbf{\textit{Contextual Reasoning} (\context)} requiring skills in the context window.
Formally, the capability requirement is a logical conjunction: $\mathcal{C}_{\text{\comp}}\iff\mathcal{C}_{\text{\para}}\land\mathcal{C}_{\text{\context}}.$
A failure in either atomic skill ($\mathcal{C}_{\text{\para}}$ or $\mathcal{C}_{\text{\context}}$) necessitates failure in \comp.
Intuitively, humans find it straightforward to tackle a \comp task if the novel information is given. However, while fluent in $\mathcal{C}_{\text{\para}}$ and $\mathcal{C}_{\text{\context}}$ by training with massive (Context-)Question-Answer data, LLMs struggle to generalize to the composite $\mathcal{C}_{\text{\comp}}$ \citep{cheng2024understanding}.

In this paper, we study \comp through \textit{multi-hop factual reasoning}. Typically, this task requires the retrieval and composition of multiple facts, \textit{i.e.,} traversing a relational path $P = (r_1, r_2, \dots, r_k)$ starting from a topic entity to the answer. For \comp, each relation $r_i$ draws from either parametric or contextual knowledge, demanding seamless integration of both sources. For example (Figure \ref{fig:main_figure}a), determining ``\textit{the occupation of the business partner of Global View's new chief editor}'' traverses ``\textit{Chief Editor -- Business Partner -- Occupation}''.

\textbf{Generalization Levels}
\label{sec:gen_level}
Standard independent and identically distributed (IID) data splits fail to capture the combinatorial complexity of real-world reasoning, \textit{e.g.,} a web-agent often encounters infinite combinations of operations that cannot be exhaustively covered in a finite training set \citep{deng2023mind2web}. To rigorously study whether models transcend rote memorization to novel scenarios, we evaluate across \textit{three levels of generalization} based on the novelty of relational path \citep{gu2021beyond}.
Let $\mathcal{P}_{\mathrm{train}}$ and $\mathcal{P}_{\mathrm{train}}$ denote the set of paths in the training and testing set, and $\mathcal{R}_{\mathrm{train}}$ denote the set of individual relations in those paths. We categorize the generalization levels as follows (Figure \ref{fig:main_figure}b):

\noindent$\bullet$ {\textbf{\textit{IID Generalization}} evaluates the application of learned patterns. The tested relational path is fully seen during training ($P_{\mathrm{test}} \in \mathcal{P}_{\mathrm{train}}$).
For instance, if trained on ``\textit{Business\_Partner}$\rightarrow$\textit{Occupation}'' (Job of [X]'s business partner), an IID test queries the same structure for a different entity. 
Experiments show that SFT is sufficient for this level as it merely requires recalling the observed structure.

\noindent $\bullet$ \textbf{\textit{Composition Generalization}} tests the ability to recombine in-distribution primitives into novel reasoning patterns.
While every individual relation constituting the path has been observed in disjoint contexts during training ($\forall r \in P_{\mathrm{test}}, r \in \mathcal{R}_{\mathrm{train}}$), the specific sequence of relations is unseen ($P_{\mathrm{test}} \notin \mathcal{P}_{\mathrm{train}}$).
For example, suppose the training set contains \textit{Business\_Partner}$\rightarrow$\textit{Job} and \textit{Business\_Partner}$\rightarrow$\textit{Spouse}. A Composition test might query  \textit{Spouse}$\rightarrow$\textit{Job}.
Although the model knows both relations independently, it must synthesize them into a new compound reasoning path without explicit prior demonstration.

\noindent$\bullet$ \textit{\textbf{Structural Zero-shot Generalization}} is the most challenging.
Unlike standard zero-shot prompting, it tests unseen relational primitives during instruction-tuning.
The path involves at least one relation never seen in any QA pair during training ($\exists r \in P_{\mathrm{test}}, r \notin \mathcal{R}_{\mathrm{train}}$).
While the model may semantically understand the relation ``\textit{Advisor}'' from pre-training, it has never been explicitly trained to retrieve or reasoning with it in a multi-hop reasoning path (\textit{i.e.}~``the spouse of [X]'s {advisor}'').

\section{Experiment Setup}
\label{sec:exp_setup}

\subsection{Semantic-Synthetic Human Biographies}
\label{sec:syn_human_bio}

To avoid data contamination and control knowledge sufficiency, we construct a synthetic biography dataset grounded in a relational Knowledge Graph (KG) \citep{allen2023physics3.1, allen2023physics3.2}.
We define 39 relations, randomly partitioned into disjoint Parametric and Contextual sets.
We populate the KG with meaningful synthetic facts (\textit{e.g.,} name, child) to guarantee no overlap with pre-training corpora.
We generate 10k biographies each for \para and \context, with 5k shared entities to enable bridging, simulating real-world scenarios where novel facts attach to known entities (Appendix \ref{app:knowledge_graph_details}).

\textbf{Reasoning Pattern Construction from KG}
\label{sec:qa_pair_construct}
To build multi-hop questions, we sample 2--5 hop relational paths distributed identically for each reasoning type. We require intermediate node to be entity rather than generic value (\textit{e.g.,} date, email) to enforce multi-step dependency. Paths in \comp strictly integrate relations from both \para and \context. We use three linguistically diverse templates to prevent syntactic shortcuts. Crucially, target outputs include \textit{Chain-of-Thought} \citep{wei2022chain}, facilitating query decomposition, aids retrieval and application of knowledge \citep{allen2023physics3.1}, and allows to track precisely whether the model retrieves facts from parameter or context.

\textbf{Data Split for Generalization Levels} 
To evaluate the levels defined in \S \ref{sec:gen_level}, we implement a rigorous splitting protocol at the \textit{relational path} level:

 \textbf{1) IID}:
A subset of relational paths $\mathcal{P}_{\textrm{IID}}$ is shared between both train and test.
To prevent rote memorization, we populate $\mathcal{P}_{\textrm{IID}}$ with disjoint entities.

 \textbf{2) Structural Zero-shot}:
We designate a subset of relations as $\mathcal{R}_{\textrm{unseen}}$.
\textit{Every} path containing a relation $r \in \mathcal{R}_{\textrm{unseen}}$ is assigned to the zero-shot test set, guaranteeing zero exposure to these unseen relations during training.

 \textbf{3) Composition}:
We randomly partition the remaining paths (\textit{i.e.,}~not IID, not Structural Zero-shot) into train set $\mathcal{P}_{\textrm{train\_comp}}$ and test set $\mathcal{P}_{\textrm{test\_comp}}$.
Crucially, we strictly enforce that the underlying relation vocabularies are identical ($\mathcal{R}_{\textrm{test}} \equiv \mathcal{R}_{\textrm{train}}$), ensuring difficulty stems solely from novel \textit{combination} of known relations.

Through this protocol, we guarantee that IID tests seen paths, Composition tests unseen paths composed of seen relations, and Zero-shot tests paths with unseen relations.
We then construct QA pairs via random-walk over the KG: sample an entity as the starting point and traverse the path to the answer, terminating with the phrase ``So, the answer is:''.
As construction is scalable, we provide sufficient training samples to ensure high IID SFT performance ($> 90\%$).
Details in Appendix \ref{app:knowledge_graph_details}.

\begin{table}[t]
    \centering
    \caption{Statistics of train and test set.}
    \label{tab:stats_a}
    \vspace{-7pt}
    \small
    \begin{tabularx}{\columnwidth}{lXXXX}
    \toprule
        \textbf{Group} & \textbf{Train~~} & \textbf{IID} & \textbf{Com.} & \makecell{\textbf{0-shot}} \\ 
    \midrule
        Parametric    & 88,031  & 1,921 & 1,141 & 782 \\
        Contextual    & 2,651   & 1,910 & 1,320 & 453 \\
        Complementary & 180,919 & 2,135 & 1,415 & 918 \\
    \bottomrule
    \end{tabularx}
    \vspace{-4pt}
\end{table}
\begin{table}[t]
    \centering
    \small
    \vspace{-6pt}
    \caption{SFT results on \comp test set.}
    \vspace{-8pt}
    \label{tab:empirical_results_b}
    \begin{tabularx}{\columnwidth}{Xccc}
    \toprule
        \textbf{Training Data} & \textbf{IID} & \textbf{Com.} & {\textbf{0-shot}} \\
    \midrule
        Parametric + Contextual & 35.18 & 28.20 & 24.07 \\
        Complementary           & 90.30 & 76.25 & 18.41 \\
    \bottomrule
    \end{tabularx}
    \vspace{-.5cm}
\end{table}

\begin{figure*}[t]
    \setlength{\abovecaptionskip}{0.1cm}
    \setlength{\belowcaptionskip}{-0.4cm}
    \centering
    \includegraphics[width=\linewidth]{images/figure1_relative_abs_new.png}
    \caption{Comparison of training with different complementary data proportions. {The top row} compares the gain from RL, while {the bottom row} compares the absolute performance after RL. It shows that LLMs generalize to Complementary Reasoning only from the model with both Parametric and Contextual Reasoning skills.}
    \label{fig:sufficiency_comparison}
\end{figure*}

\subsection{Training Protocol}
\label{sec:training_protocol}

We adopt a standard SFT-then-RL pipeline using {Qwen-2.5-1.5B} (More in Appendix \ref{app:robustness_of_findings}).
\textbf{Input Formulation}: \para inputs consist solely of the \textit{Question}, whereas \context and \comp integrate both the \textit{Context} (novel facts) and the \textit{Question}.
\textbf{Training}: SFT uses standard next-token prediction on QA pairs and parametric biographies.
RL uses Group Relative Policy Optimization (GRPO) \citep{shao2024deepseekmath} with binary outcome rewards. Evaluation uses exact matching on the \comp test set.

\subsection{Baseline Analysis: The Limits of SFT}
\label{sec:baseline_analysis}

To establish why SFT alone is insufficient for robust Complementary Reasoning, we analyze task difficulty and evaluate two SFT baselines: training with atomic skills (\sftatomic) and training directly with the target composite task (\sftcomp).


\textbf{\comp is structurally complex and data-hungry.} 
Table~\ref{tab:stats_a} shows that achieving high IID performance on \comp requires significantly more training data (180k) than the atomic \para (88k) and \context (3k) tasks.
This validates that integrating internal and external knowledge demands more than simply applying skills in isolation.

\textbf{Atomic skills do not spontaneously compose.} 
While \sftatomic learns the necessary foundational components (\para and \context), it exhibits poor transfer to the composite skill (\comp) (Table~\ref{tab:empirical_results_b}).
It lags far behind the explicit training baseline (90.26\% IID), confirming that possessing atomic skills does not guarantee their successful integration without further guidance.

\textbf{SFT Memorizes rather than Generalizes.} 
While \sftcomp achieves a near-perfect 90\% in IID, it collapses to 18\% in Zero-shot (Table~\ref{tab:empirical_results_b}).
This indicates that SFT incentivizes memorizing relational patterns seen during training; but fails when faced with novel combinations where memorization is impossible.

\begin{figure*}
    \centering
    \setlength{\abovecaptionskip}{0.cm}
    \setlength{\belowcaptionskip}{-0.3cm}
    \includegraphics[width=\linewidth]{images/four_comparison_bars_new.png}
    \caption{Necessity of atomic skills for RL generalization. We conduct RL with the same amount of \comp data from SFT models with different atomic foundations. Only \sftatomic generalizes well in all levels.}
    \label{fig:necessity_comparison}
\end{figure*}

\section{RL Enables Generalization}


We hypothesize that generalization requires a specific curriculum: \textit{establishing sufficient atomic abilities via SFT, followed by synthesizing complex skills via RL}. We propose the recipe \sftatomic$\rightarrow$\rlcomp as the optimal path to generalization. Crucially, as we will demonstrate via \textit{pass@$k$} analysis (\S \ref{sec:role_of_RL}), this specific curriculum is what enables RL to act as a genuine \textit{synthesizer} of new capabilities, rather than a mere probability amplifier. We validate the efficacy of this recipe by testing both its \textbf{sufficiency} (consistent performance gains) and \textbf{necessity} (data and training prerequisites).

\subsection{Sufficiency: Atomic-First Beats Composite-First at Every Data Scale}\label{sec:sufficiency}


Our controlled experiments reveal that, regardless of data scale, our recipe exceeds direct composite training, and the gap widens sharply on out-of-distribution settings. Figure~\ref{fig:sufficiency_comparison} visualizes this dominance; the remainder of this subsection breaks down why. More in Appendix \ref{app:sufficiency}.

\noindent\textbf{Settings.} While combining an SFT cold-start with extensive RL is standard practice \citep{guo2025deepseekr1}, the optimal data balance remains unclear.
To study this, we partition the \comp training set into an SFT subset ($x\%$) and an RL subset ($(100-x)\%$), varying $x$ from 10 to 90, and compare two recipes:
\textbf{1)} \textbf{\sftcomp$\rightarrow$\rlcomp}: SFT on the $x\%$ of the composite data, then RL on the remaining $(100-x)\%$;
\textbf{2)} \textbf{\sftatomic$\rightarrow$\rlcomp}: SFT on all atomic data (\para$+$\context, roughly equal in volume to $50\%$ of \comp data), followed by \textit{RL on the same $(100-x)\%$ \comp subset}.
Figure~\ref{fig:sufficiency_comparison} reports relative performance gain from RL (top) and final absolute performance (bottom).

\noindent\textbf{RL efficiently synthesizes atomic skills.}
The top row of Figure~\ref{fig:sufficiency_comparison} reveals a striking contrast.
Across all data volumes (10-90\%), \sftatomic (red bar) achieves massive performance gains during the RL phase on \comp data.
Conversely, \sftcomp (green bar) gains almost nothing from RL.
This indicates that when directly trained on composite tasks without atomic foundations, RL merely optimizes within an existing distribution.
However, when grounded in atomic skills, RL acts as a powerful synthesizer, actively exploring and composing known facts into new reasoning skills.

\noindent\textbf{Superior Zero-shot generalization of RL over atomic skills.}
In the critical Structural Zero-shot setting (3rd Column) where rote memorization is impossible, \sftatomic (yellow bar) consistently and significantly outperforms \sftcomp (blue bar) regardless of data scale.
This confirms that our recipe fosters genuine generalization to unseen relations, whereas direct training on composite data fails to extrapolate beyond the training distribution.

\noindent\textbf{The SFT Generalization Paradox.}
We observe a nuanced ``memorization trap'' in the IID and Composition settings (Bottom Row, 1st \& 2nd Columns).
As the data ratio $x$ exceeds 70\%, the absolute IID performance of \sftcomp (blue bar) slightly surpasses \sftatomic (yellow bar), yet it still collapses in Zero-shot. This indicates that excessive SFT on composite data encourages overfitting to specific training relation paths rather than learning the underlying reasoning algorithm. In contrast, RL built upon atomic skills avoids this trap by forcing the model to dynamically derive paths, resulting in robust OOD performance.

 

\subsection{Necessity of Atomic Skill (Data Condition)} \label{sec:necessity_data}

To determine whether \textit{\textbf{complete atomic capabilities are strictly prerequisite for generalization}}, we compare \sftatomic against baselines with similar initial performance but deficient atomic foundations: \textbf{1)} $\textsc{SFT}_{\para}$ (\para only); \textbf{2)} $\textsc{SFT}_{\context}$ (\context only); \textbf{3)} $\textsc{SFT}_{10\% \comp}$ and $\textsc{SFT}_{20\% \comp}$ (SFT on \comp data with comparable initial performance). We apply identical RL (a subset of 12.8k \comp samples) to every model (Figure~\ref{fig:necessity_comparison}).



\textbf{Removing any atomic skill collapses generalization.} Models trained via SFT solely on \para or \context fail to generalize significantly after RL, proving that Complementary Reasoning is not merely an additive task but a synthesis requiring full atomic sufficiency. 

\textbf{Generalization potential is driven by foundational capability, not initial task metrics.}
While $\textsc{SFT}_{10\% \comp}$ and $\textsc{SFT}_{20\% \comp}$ match \sftatomic's pre-RL performance, their post-RL gain is negligible.
Conversely, \sftatomic nearly doubles its performance across all levels.
Strikingly, even initially low-performing $\textsc{SFT}_{\para}$ and $\textsc{SFT}_{\context}$ gain more from RL than \sftcomp does, confirming that underlying atomic skills, not initial performance, are the true predictor of RL success.

\begin{figure}
\vspace{-0.35cm}
    \setlength{\abovecaptionskip}{-0.1cm}
    \setlength{\belowcaptionskip}{-0.3cm}
    \includegraphics[width=\linewidth]{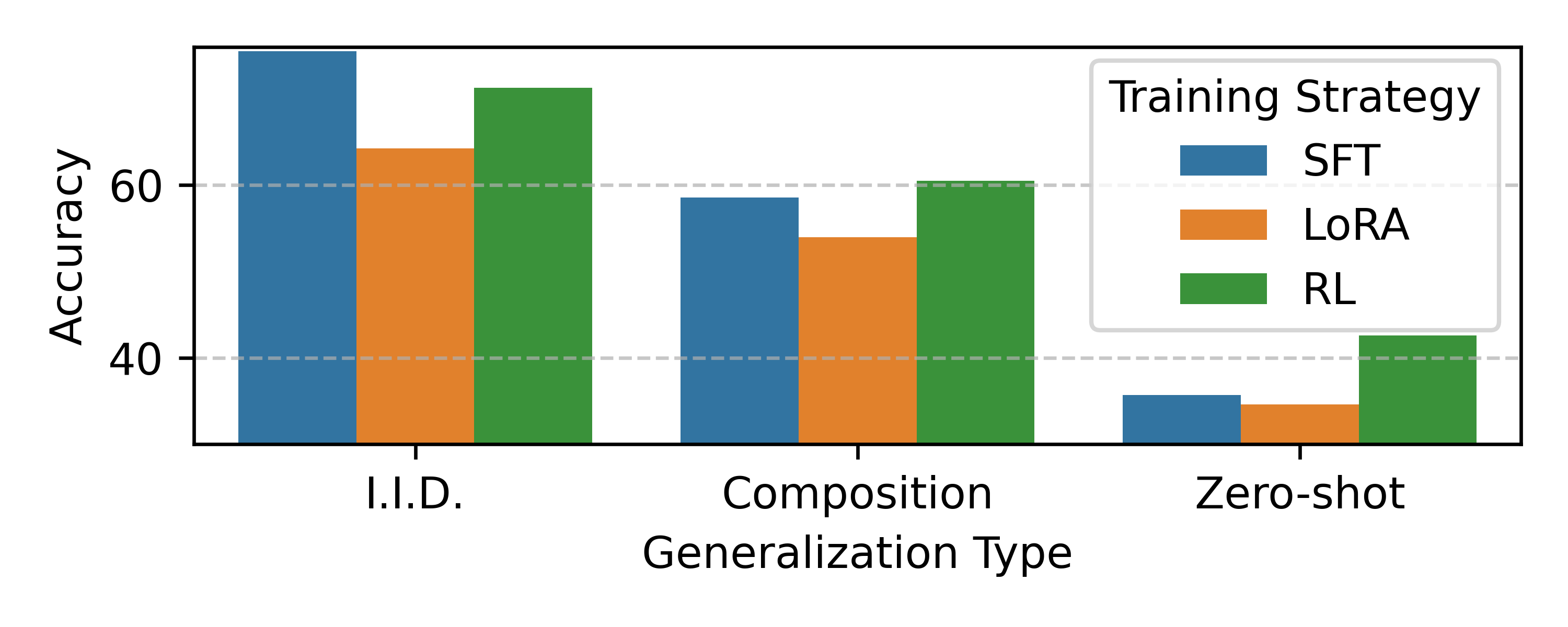}
    \caption{Performance of different training strategies over 12.8k \comp samples across  generalization levels.}
    \label{fig:training_strategy}
    \vspace{-4pt}
\end{figure}

\begin{figure*}
    \centering
    \setlength{\abovecaptionskip}{0.1cm}
    \setlength{\belowcaptionskip}{-0.3cm}
    \includegraphics[width=\linewidth]{images/multi_gen_sft_rl_performance_comparison_new.png}
    \caption{Performance with the same amount of SFT and RL data, comparing \sftatomic and \sftcomp.}
    \label{fig:sample_eff_huge}
\end{figure*}

\subsection{Necessity of RL (Training Condition)}
\label{sec:necessity_training}

Given sufficient atomic skills (\sftatomic), we now ask \textbf{whether RL uniquely is needed, or whether further SFT would suffice.}
We compare training strategies (SFT, LoRA [rank=256], and RL) using identical \comp samples
(Figure \ref{fig:training_strategy}). A clear dichotomy emerges between memorization and generalization: further SFT (blue) yields the highest IID performance (indicating memorization of seen patterns) but severely lags in {Zero-shot}, while RL (green) more than doubles SFT/LoRA performance on unseen combinations. While SFT excels at pattern matching, \textbf{RL is uniquely necessary to incentivize the active composition of skills required for OOD reasoning.}

\section{Why Atomic-First Works: Analyses}
\label{sec:analysis}

Having established atomic skills as a prerequisite for RL generalization, we analyze the features of our atomic-first recipe from sample efficiency, pass@$k$ performance and behavioral case study.
More analyses (training dynamics, loss impacts, latent space PCA study, longer-hop extrapolation and uncertainty analysis) in Appendix \ref{app:other_factors}.

\subsection{Sample Efficiency}

We examine \textbf{whether learning atomic skills induces better sample efficiency than learning the composite task directly}, both before and after RL.

\textbf{Atomic learning requires less SFT data to prime RL generalization.}
To compare sample efficiency, we reserve a fixed subset ($\sim$90k) of \comp data exclusively for the RL phase.
For the SFT phase, we sweep a data budget (20\%--100\% of the remaining volume) to construct two strictly comparable datasets:
\textbf{1): composite samples} (\sftcomp);
\textbf{2) atomic samples} (\sftatomic).
To ensure fairness, we control both the \textit{number of samples} and  \textit{information content} (strictly matching the distribution of reasoning steps--hop counts across the two), ensuring that models face equivalent reasoning complexities.
We then apply RL on the reserved set.
Figure~\ref{fig:sample_eff_huge} shows that \sftatomic (red/orange) consistently outperforms \sftcomp (green/blue) across all SFT data scales.
Even with a minimal SFT budget ($\sim$18k), \sftatomic is successfully ``primed'' for RL, whereas \sftcomp struggles. This confirms that atomic skills provide a highly efficient foundation for complex reasoning.

\textbf{Atomic skills enable few-shot adaptation.}
We next examine the volume of \comp data required to ``trigger'' generalization once sufficient atomic skills are established.
Fixing \sftatomic as base model, we apply RL, SFT, or LoRA with data scaling from a tiny (50 samples) to a medium set (12.8k, $<10\%$ of total) and compare against an upper-bound baseline: SFT on 100\% \comp data.
Figure~\ref{fig:sample_efficiency} reveals \textbf{rapid adaptation:} with just 50 samples, \sftatomic significantly improves in complementary reasoning regardless of tuning strategy. We also observe extreme \textbf{data efficiency:} with $<$10\% of the composite data, \sftatomic effectively matches the upper-bound of \sftcomp trained on the entire dataset (purple dotted line).
This demonstrates that once atomic skills are acquired, the cost of assembling them into a complex reasoning skill is remarkably low.

\begin{figure}
    \vspace{-0.4cm}
    \centering
    \setlength{\abovecaptionskip}{-0.1cm}
    \setlength{\belowcaptionskip}{-0.4cm}
    \includegraphics[width=\linewidth]{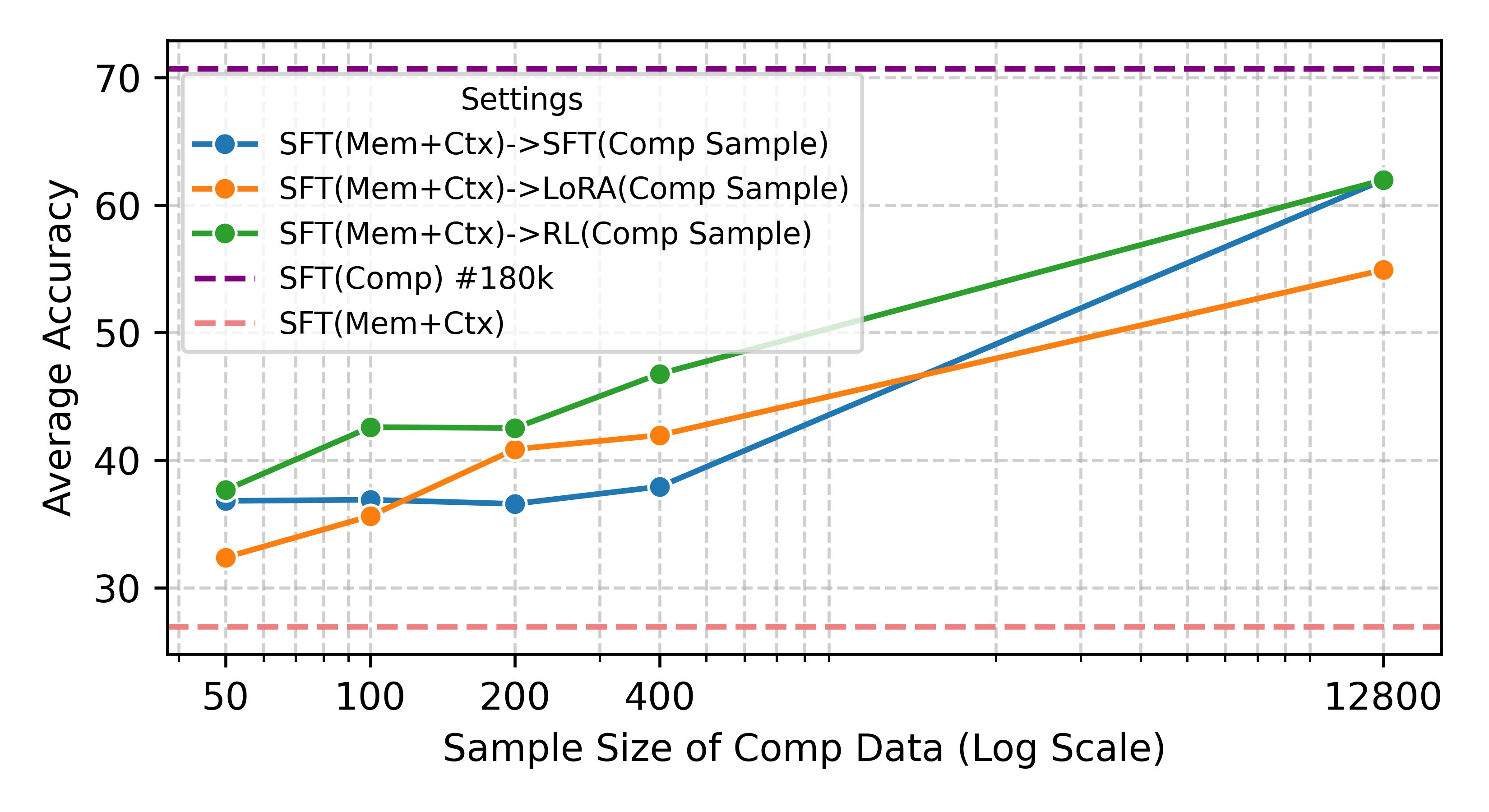}
    \caption{Few-shot adaptation of \sftatomic. We show average accuracy over all generalization levels.}
    \label{fig:sample_efficiency}
\end{figure}

\begin{figure*}
    \centering
    \setlength{\abovecaptionskip}{0.cm}
    \setlength{\belowcaptionskip}{-0.3cm}
  \includegraphics[width=\linewidth]{images/passk_comparison_combined_figure_new.png}
    \caption{Pass@$k$ comparison for \sftatomic and \sftcomp. RL synthesizes new compositional skills only based on models with sufficient atomic skills.}
    \label{fig:pass@k}
\end{figure*}

\subsection{The Role of RL: Synthesizer vs. Amplifier}
\label{sec:role_of_RL}

As telegraphed earlier, we analyze \textit{pass@$k$} performance before and after RL \citep{yue2025does}, varying $k$ from $2^0$ to $2^9$, to determine if RL acts as a \textbf{synthesizer} (divergent curves at large $k$, implying novel skills discovered) or an \textbf{amplifier} (converging curves, implying skills already existed in the SFT distribution). We compare \sftatomic and \sftcomp using identical RL data (Figure~\ref{fig:pass@k})


\textbf{RL synthesizes new pathways given sufficient atomic skills.} 
\sftatomic (top row) shows parallel scaling: RL performance (orange) remains significantly higher than SFT (blue) even at maximal $k=2^9$. If RL were merely amplifying latent behaviors, the SFT model (given enough attempts) would eventually sample the correct reasoning path.
The persistent gap of pass@$k$ curves suggests a fundamental mechanistic shift: RL has synthesized a robust, novel mechanism---\textit{specifically, bridging context and memory}---that is functionally absent from the SFT distribution.
This provides empirical evidence that, given sufficient atomic priors, RL creates new capabilities rather than just re-weighting existing ones.
Notably, the discovery of new reasoning mechanisms occurs across all generalization levels, suggesting that RL optimizes the logic of combination itself.

\textbf{Without sufficient atomic skills, RL merely amplifies.}
Conversely, for \sftcomp (bottom row), the \textit{pass@$k$} curves rapidly converge as $k$ increases.
By $k=2^5$, the SFT model effectively matches the RL model.
Because \sftcomp memorized the composite target distribution during SFT, RL serves only as an amplifier---boosting the likelihood of the optimal reasoning paths without discovering fundamentally new ones.
This dichotomy confirms our core thesis: atomic sufficiency is the strict prerequisite for RL to function as a synthesizer; otherwise, RL degenerates into a mere probability amplifier.

\subsection{Where RL Shifts the Failure Mode}

\S \ref{sec:role_of_RL} shows {that} atomic-first RL synthesizes new pathways. We now ask \textit{what} those pathways look like behaviorally. We analyze failure modes by examining the \textbf{intersection of incorrect samples} before and after RL to identify how RL changes detailed reasoning behavior. 
For persistent errors, we align the generated CoT against the ground truth to pinpoint the first deviating step, classify the error source (\para vs.\ \context), and compute \textbf{Progress} (the normalized position of the failure step within the reasoning path). Table~\ref{tab:error_analysis} summarizes the patterns.

\textbf{Without atomic-grounded RL, failures cluster at early contextual steps.}
\sftatomic, \sftcomp, and \sftcomp$\rightarrow$\rlcomp share a failure profile, characterized by high \context error ($>85\%$) and early-stage failures (Progress $<55\%$). Qualitatively, \sftatomic frequently hallucinates when retrieving from the provided context, similar to its imitation-learning origin in contextual reasoning. Conversely, \sftcomp, struggles to identify the correct relation when bridging from \para to \context, with 62\% of its errors terminating at the very first hop. Even adding RL on \sftcomp barely changes this profile, mirroring that RL without atomic priors only amplifies the existing distribution.

\textbf{RL shifts the bottleneck to late-stage parametric recall.} 
In sharp contrast, \textbf{{\sftatomic$\rightarrow$\rlcomp}} flips this distribution: 70\% of errors are from \para, and failures occur significantly later (71.8\% Progress), typically at the final hop.
This indicates that \textit{RL successfully optimizes the bridging logic, pushing failure modes from early contextual retrieval to late parametric recall}.

\begin{table}[t]
    \centering
    \vspace{-0.2cm}
    \small
    \caption{Error analysis. \textbf{{\para}} and \textbf{{\context}} denotes error occurring at parametric and contextual relations, respectively. \textbf{{Prgs}} denotes normalized position of the first error within the reasoning path.}
    \vspace{-0.2cm}
    \begin{tabularx}{\columnwidth}{lXXX}
        \toprule
        \textbf{Model} & \textbf{\context} & \textbf{\para} & \textbf{Prgs} \\
        \midrule
         \sftcomp & 90\% & 10\% & 54.5\% \\
        \sftcomp$\rightarrow$\rlcomp& 86\% & 14\% & 45.0\% \\
       \midrule
       \sftatomic & 86\% & 14\% & 18.5\% \\
        \sftatomic$\rightarrow$\rlcomp & 30\% & 70\% & 71.8\% \\
        \bottomrule
    \end{tabularx}
    \label{tab:error_analysis}
    \vspace{-0.4cm}
\end{table}

\section{Conclusions}

We investigate RL-driven generalization through \textbf{Complementary Reasoning}. 
Our findings resolve the debate on whether RL synthesizes new skills or merely amplifies existing ones: it does \textit{both}, depending entirely on the model's foundation.
We uncovered the \textbf{SFT Generalization Paradox},
showing that direct supervision on complex tasks incentivizes the memorization of specific reasoning paths, leading to brittle OOD performance.
However, we demonstrate that RL is the key to breaking this memorization trap, but only under a strict prerequisite of \textbf{sufficient atomic skills}.
When the condition is met, RL actively synthesizes novel reasoning strategies---bridging parametric and contextual knowledge---absent in the SFT distribution.
This prescribes a clear, scalable recipe for post-training: rather than collecting expensive traces of complex reasoning, one should prioritize the efficient acquisition of atomic skills via SFT, then leverage RL to catalyze the compositional generalization.

\section*{Limitations}
Our work establishes the \sftatomic$\rightarrow$\rlcomp recipe as a principled path to compositional generalization through controlled, contamination-free experimentation. It opens several promising directions. Our multi-angle behavioral analyses (pass@$k$, sample efficiency, training dynamics, latent-space PCA, and error patterns) characterize \textit{that} atomic-grounded RL synthesizes new compositional skills; a natural next step can be localizing how RL recruits and recombines the attention heads responsible for atomic parametric and contextual retrieval. Relatedly, our use of synthetic biographies is the design choice that allows us to strictly enforce the boundary between parametric and contextual knowledge---a boundary that is otherwise blurred by pre-training contamination---and validating the recipe on naturalistic benchmarks under controlled contamination conditions (\textit{e.g.,} news-QA splits dated strictly post pre-training cutoff) would confirm transfer to deployed RAG settings. Finally, while we isolate our contributions of atomic SFT and composite RL to make their roles legible, studying how to jointly mix \comp, \para, and \context data---and how the mixing schedule should evolve across SFT and RL phases---is a promising direction for practitioners adopting this recipe at scale.

\bibliography{custom}
\newpage
\appendix

\begin{figure*}
    \centering
    \includegraphics[width=\linewidth]{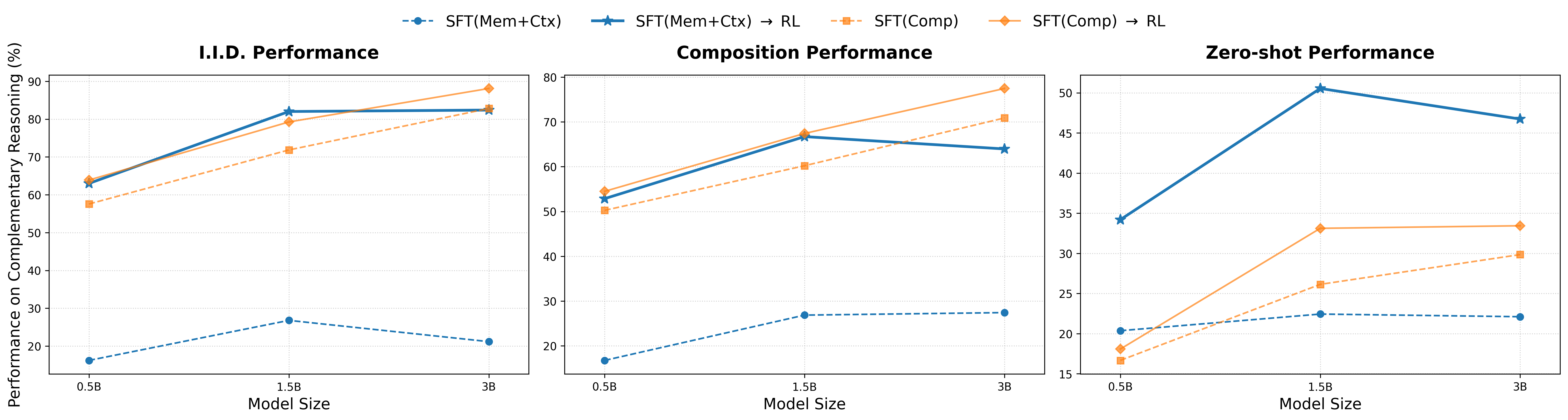}
    \caption{Model scaling analysis of Qwen model family across 0.5B, 1.5B, and 3B parameters.}
    \label{fig:model_scaling}
\end{figure*}


\section{Data and Setups}

We will release the data upon publication.

\subsection{Data Construction Details}
\label{app:knowledge_graph_details}

To systematically construct our synthetic human biographies, we build a relational knowledge graph with fake information by Python Faker Library \footnote{https://faker.readthedocs.io/en/master/} and the help of GPT-4o.

We finally synthesize 39 relations including eight symmetric relations (\textit{e.g., spouse, sibling}), and eight pairs of inverse relations (\textit{e.g., child and parent}) to mimic real-world complexity. For each relation, we adopt an LLM to construct and validate three natural language templates. We adopt heuristic rules to constrain the tailed entity of the relation (\textit{e.g.,} birthday should be a date). Table \ref{tab:relation_templates} shows the relations and templates.

For QA pairs construction, after sampling the relation paths/combinations, we adopt GPT-4o to generate question templates. Table \ref{tab:template_question_example} shows an example question template. The QA setting is the same as the standard retrieval-augmented generation (RAG) systems \cite{gutierrez2025rag,an2025thread,zhuang-etal-2024-efficientrag,cheng2024call}

After we split the relation combinations based on the generalization levels \ref{sec:gen_level}, we can synthesize human biographies based on the knowledge graph. We firstly conduct random walk over the knowledge graph entities, and then translate the obtained dict-formed biographies into natural language paragraphs with the relation templates.  Table \ref{tab:human_biography} shows an example of the biography dict and corresponding natural language paragraph.

\begin{table}[t]
    \centering
    \small
    \caption{Empirical SFT performance of three reasoning types, trained on the training set and tested on the corresponding test set. Note that this is different from other experiments that focus on evaluation over complementary reasoning test set.}
    \label{tab:Training_performance_base}
    \begin{tabular}{lccccc}
    \toprule
       \textbf{Training Data}  &  \textbf{IID} & \textbf{Comp} &  \textbf{Zero-shot}  \\
       \midrule
       Parametric Reasoning  &  93.96 &	82.30 &	3.71  \\
       Contextual Reasoning &  98.53 & 95.53 & 69.53  \\
       Complementary Reasoning & 90.26 & 76.61 & 7.76 \\
       \bottomrule
    \end{tabular}
\end{table}

\begin{table*}[t]
\centering
\caption{An example of Question and CoT answer templates based on a relation path (4-hop). \{e1\} denotes the topic entity in the question.}
\label{tab:template_question_example}
\small 
\renewcommand{\arraystretch}{1.3} 
\begin{tabular}{l p{10cm}} 
\toprule
\textbf{Component} & \textbf{Content / Examples} \\ 
\midrule
\textbf{Relation Path} & $\texttt{sibling} \xrightarrow{} \texttt{boss\_of} \xrightarrow{} \texttt{mentored\_by} \xrightarrow{} \texttt{best\_friend}$ \\ 
\midrule
\multirow{6}{*}{\textbf{Question Template}} 
 & 1. Who is the best friend of the person mentoring the employee of the sibling of \{e1\}? \\
 & 2. Can you tell me the best friend of the person who mentored the employee of \{e1\}'s sibling? \\
 & 3. Who is the best friend of the person mentoring the employee of the sibling that \{e1\} has? \\ 
\midrule
\multirow{6}{*}{\textbf{CoT Answer Template}} 
 & 1. \{e2\} is \{e1\}'s brother/sister. \{e3\} works under \{e2\}. \{e3\} was trained by \{e4\}. \{e4\}'s best friend is \{e5\}. So, the answer is: \{e5\} \\
 \vspace{1mm} 
 & 2. \{e1\} and \{e2\} are siblings. \{e2\} is the boss of \{e3\}. \{e3\} was trained by \{e4\}. \{e4\}'s best friend is \{e5\}. So, the answer is: \{e5\} \\
 \vspace{1mm}
 & 3. \{e2\} is \{e1\}'s brother/sister. \{e2\} is the boss of \{e3\}. \{e3\} received guidance from \{e4\}. \{e5\} is \{e4\}'s closest friend. So, the answer is: \{e5\} \\ 
\bottomrule
\end{tabular}
\end{table*}

\begin{table*}[t]
\centering
\small
\caption{Replication results using \texttt{Llama-3.2-1B} compared with \texttt{Qwen-2.5-1.5B}. The results confirm that our recipe facilitates significantly better zero-shot generalization, while composite SFT favors more in IID and composition settings.}
\label{tab:llama_reproduction}
\begin{tabular}{llccc}
\toprule
\textbf{Model} & \textbf{Setting} & \textbf{IID} & \textbf{Composition} & \textbf{Zero-shot} \\
\midrule
\multirow{4}{*}{\textbf{Qwen-2.5-1.5B}} & \sftatomic & 26.79 & 26.86 & 22.44 \\
 & \sftcomp & 71.85 & 60.21 & 26.14 \\
 & \sftcomp $\rightarrow$ \rlcomp & 79.25 & 67.42 & 33.12 \\
& \sftatomic $\rightarrow$ \rlcomp & {82.01} & {66.71} & {50.54} \\
\midrule
\multirow{4}{*}{\textbf{Llama-3.2-1B}} & \sftatomic & 14.85 & 20.42 & 20.81 \\
 & \sftcomp & 51.33 & 40.95 & 10.46 \\
 & \sftcomp $\rightarrow$ \rlcomp & 83.75 & 71.52 & 17.10 \\
  & \sftatomic $\rightarrow$ \rlcomp & {74.85} & {58.80} & {36.93} \\
\bottomrule
\end{tabular}
\end{table*}

\begin{table*}[t]
\centering
\small
\caption{Performance comparison across reasoning hop counts. Our recipe shows superior extrapolation to longer chains, particularly in the Zero-shot setting where 4- and 5-hop tasks are rare during training.}
\label{tab:hop_generalization}
\begin{tabular}{llcccc}
\toprule
\textbf{Setup} & \textbf{Level} & \textbf{2-Hop} & \textbf{3-Hop} & \textbf{4-Hop} & \textbf{5-Hop} \\ \midrule
\sftcomp$\rightarrow$\rlcomp & Composition & 77.64 & 61.90 & 50.50 & 46.67 \\
& Zero-shot & 25.00 & 44.00 & 35.64 & 15.19 \\ \midrule
{\sftatomic$\rightarrow$\rlcomp} & Composition & 75.21 & 56.65 & 61.88 & 66.67 \\
& {Zero-shot} & {59.78} & {49.33} & {47.87} & {30.38} \\ \bottomrule
\end{tabular}
\end{table*}

\subsection{Details of SFT Training Baselines}
\label{app:SFT_training_details}

As discussed in \S \ref{sec:qa_pair_construct}, we are able to synthesize as much data as needed. However, in real-world scenarios, while LLMs can easily handle either Parametric or Contextual Reasoning, probably through post-training with sufficient data, they struggles in Complementary Reasoning task \cite{cheng2024understanding}, where it is hard to collect ample data. We show in Table \ref{tab:stats_a} that Complementary Reasoning is data-hungry and most difficult compared to parametric and contextual reasoning. Here we show empirical results training with SFT and evaluating on the corresponding testing set in Table \ref{tab:Training_performance_base}. 

It further shows that Contextual Reasoning is the easiest. While LLMs learns to adopt new knowledge during SFT training, they manage to handle most of the unseen knowledge in the context window. We hypothesize that this ability is essential for further generalization to new reasoning patterns. Moreover, SFT training involving parametric knowledge would be hard to generalize in Zero-shot settings. Both parametric and complementary reasoning, while performing well in IID setting, drop significantly in Composition and almost fail in Zero-shot setting \cite{gu2021beyond,huang2023markqa,cheng2025differentiable}.

This further highlights our motivations: we try to figure out the recipe of training strategies and mix of training data required for generalization to complementary reasoning on all difficulty levels.

\subsection{Templates and Examples of KG Relations and Biographies}

Table \ref{tab:template_question_example} shows an example of question answering template. 
Table \ref{tab:relation_templates} shows the relations and templates. Table \ref{tab:human_biography} shows an example of the biography dict and corresponding natural language paragraph.

\begin{table*}[]
\centering
\small
\caption{Statistical significance analysis using bootstrapping (1k iterations). The non-overlapping confidence intervals in the Zero-shot setting highlight the robust superiority of the atomic skill recipe.}
\label{tab:statistical_significance}
\begin{tabular}{llccc}
\toprule
\textbf{Setup} & \textbf{Level} & \textbf{Mean Acc.} & \textbf{Std.} & \textbf{95\% CI} \\ \midrule
\sftcomp$\rightarrow$\rlcomp & IID & 80.30 & 1.16 & [77.41, 83.19] \\
& Composition & 67.47 & 1.27 & [64.31, 70.63] \\
& {Zero-shot} & 31.52 & 3.55 & [22.70, 40.34] \\
\midrule
{\sftatomic$\rightarrow$\rlcomp} & IID & 83.12 & 0.97 & [80.72, 85.52] \\
& Composition & 66.31 & 0.47 & [65.15, 67.47] \\
& Zero-shot & 48.80 & 1.6  & [44.83, 52.77] \\  
\bottomrule
\end{tabular}
\end{table*}

\section{Robustness of Findings}
\label{app:robustness_of_findings}

\subsection{Model Scaling}
\label{app:model_scaling}

We study whether our findings that ``RL enables generalization in Complementary Reasoning from sufficient atomic skills''
persist as models scale. We compare Qwen-2.5-0.5B, Qwen-2.5-1.5B and Qwen-2.5-3B due to limited compute. Figure~\ref{fig:model_scaling} illustrates the performance trends across model sizes. 
We observe consistent behaviors that strongly support the effectiveness of our proposed recipe. 

As shown by the blue lines, \sftatomic leads to substantial performance jumps after \rlcomp. This improvement is particularly pronounced in the Zero-shot setting, indicating that the model with sufficient atomic skills successfully generalizes the reasoning capabilities acquired during RL. In contrast, \sftcomp (orange dashed line), while starting with decent performance, exhibits limited growth after RL training (orange solid line). The gap between the pre-RL and post-RL performance is marginal, suggesting that SFT with composite data may limit the model's potential for further generalization.
    
Crucially, \textit{these trends hold true across all model sizes}. When the model size increases to 3B, the superiority of \textit{\sftatomic$\rightarrow$\rlcomp} over \textit{\sftcomp$\rightarrow$\rlcomp} remains significant (\textit{e.g.,} a gap of approximately 13\% in Zero-shot accuracy for the 3B model). This confirms that our conclusions are robust to model scaling and implies that \sftatomic is a more effective foundation for scaling up RL-based reasoning.

\subsection{Model Diversity and Generalization (Llama)}
\label{app:model_diversity}

To further validate the robustness of our findings across different model architectures, we extend our experiments to the \texttt{Llama-3.2-1B} model. This allows us to ensure that the ``memorization trap'' of SFT and the success of the sufficient atomic recipe for RL-based skill composition are not idiosyncratic to the Qwen family.

Table~\ref{tab:llama_reproduction} shows the results of \texttt{Llama-3.2-1B}, strongly replicating our findings. Specifically, while \sftcomp achieves a baseline level of performance, it struggles significantly in the Zero-shot setting ($10.46\%$), suggesting a reliance on memorized patterns. In contrast, our proposed recipe—combining atomic skill initialization (\sftatomic) with RL—enables the model to successfully compose these skills, reaching $36.93\%$ in the Zero-shot setting.

Consistent with Appendix \ref{fig:model_scaling}, the gap in Zero-shot performance is particularly telling. The transition from \sftatomic to \rlcomp yields a substantial jump in reasoning capabilities, reinforcing the conclusion that providing models with sufficient atomic skills is a more effective foundation for RL-driven generalization than direct SFT on complex composite data. These consistent behaviors across both Qwen and Llama architectures underline the generalizability of our proposed training recipe.

\begin{figure*}
    \centering
    \includegraphics[width=0.8\linewidth]{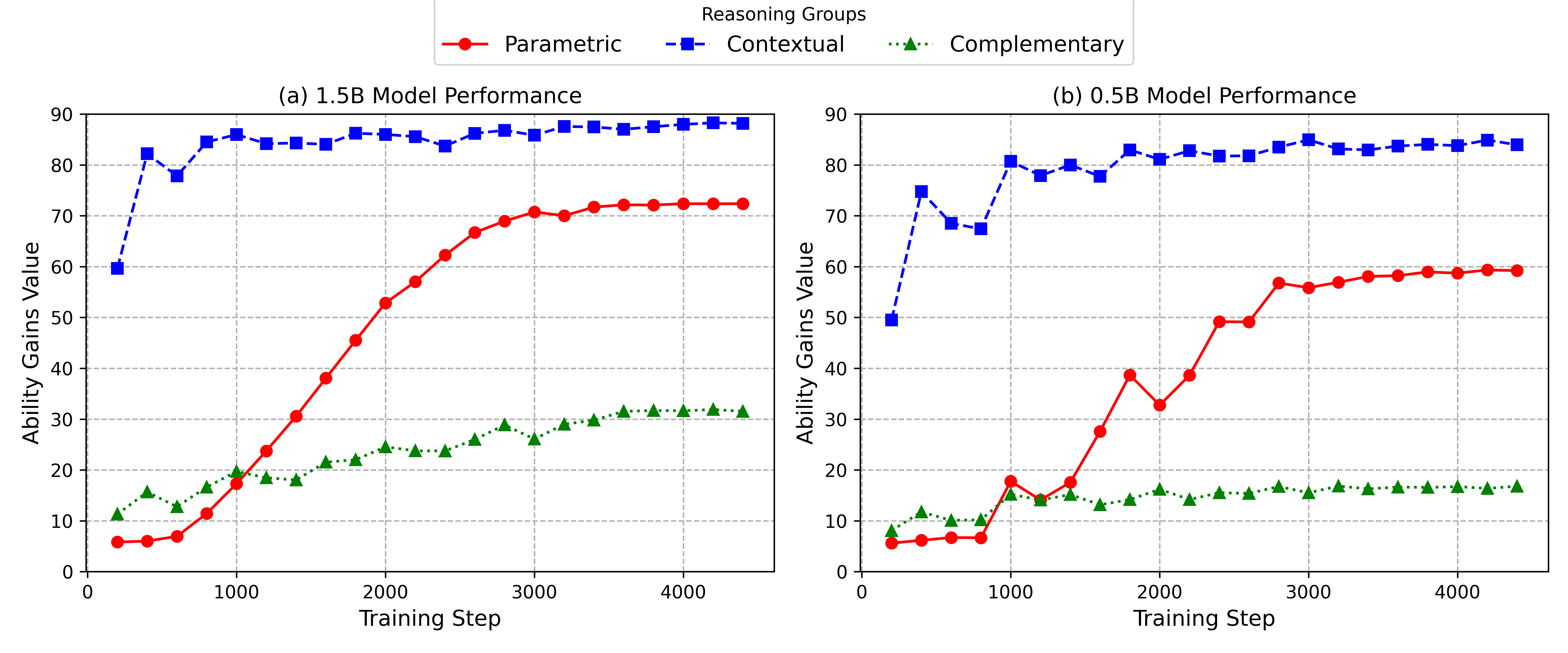}
    \caption{Training dynamics of SFT with parametric and contextual reasoning data over training steps. Ability gains are calculated over \para (Red Line), \context (Blue Line) and \comp (Green Line), respectively. As the training progresses, complementary reasoning ability emerges to some extent.}
    \label{fig:training_dynamic}
\end{figure*}

\subsection{Longer Hop Generalization}

To further evaluate the model's ability to extrapolate its reasoning capabilities, we analyze performance across varying chain lengths (number of hops). This dimension is crucial for verifying whether the model has truly acquired a generalizable reasoning mechanism or is merely relying on length-specific patterns learned during training.

\paragraph{Extrapolation to Longer Chains} 
Our training distribution is naturally skewed to favor shorter reasoning paths, with a distribution of approximately 50\%, 30\%, 15\%, and 5\% for 2, 3, 4, and 5-hop questions, respectively. Due to the scarcity of 4- and 5-hop chains in the training set (comprising less than 20\% of the data), performance on these levels serves as an effective proxy for length extrapolation. 

As shown in Table~\ref{tab:hop_generalization}, while performance naturally trends downward as the reasoning complexity increases, our proposed \textit{SFT(Mem+Ctx) $\rightarrow$ RL(Comp)} recipe exhibits a significantly more graceful degradation compared to the composite SFT baseline. Notably, in the Zero-shot setting, our recipe maintains a 30.38\% accuracy on 5-hop questions, nearly doubling the performance of the \textit{SFT(Comp) $\rightarrow$ RL(Comp)} approach (15.19\%).

\paragraph{Structural Defense Against Memorization} 
Beyond length extrapolation, the design of our data splits inherently precludes memorization. We strictly guarantee knowledge sufficiency for all cases, ensuring that the evaluation isolates the model's ability to actively synthesize and apply knowledge. In our Zero-shot setting, the evaluation paths contain at least one relation that is \textit{entirely absent} from the training set. This structural constraint makes memorization impossible, as the model must synthesize its atomic understanding of these unseen relations on the fly to successfully resolve the reasoning chain.

\subsection{Statistical Significance}

To ensure that the performance improvements observed in our experiments are statistically robust against the variance inherent in training, we evaluate our recipe across three independent training runs using different random seeds. All models are trained under identical hardware (2*H200) and other training configurations, executing the exact same number of optimization steps.

Table~\ref{tab:statistical_significance} presents the mean accuracy, standard deviation and 95\% confidence intervals (CI, calculated using the Student's t-distribution) for both the baseline (the common practice \sftcomp$\rightarrow$\rlcomp) and our proposed recipe (\sftatomic$\rightarrow$\rlcomp). While the variance across training runs results in overlapping intervals for the Compositional setting, the performance advantages of \textit{\sftatomic $\rightarrow$ \rlcomp} over \textit{\sftcomp $\rightarrow$ \rlcomp} remain distinctly robust in the IID and Zero-shot evaluations.

Of particular note is the Zero-shot setting, where the confidence intervals strictly do not overlap (\textit{e.g.,} [22.70, 40.34] vs. [44.83, 52.77]). This clear statistical separation provides strong empirical evidence that our ``atomic-first recipe'' consistently enables superior generalization in Complementary Reasoning task compared to standard composite training, independent of seed selection or initialization noise.

\begin{figure*}
    \centering
    \includegraphics[width=0.9\linewidth]{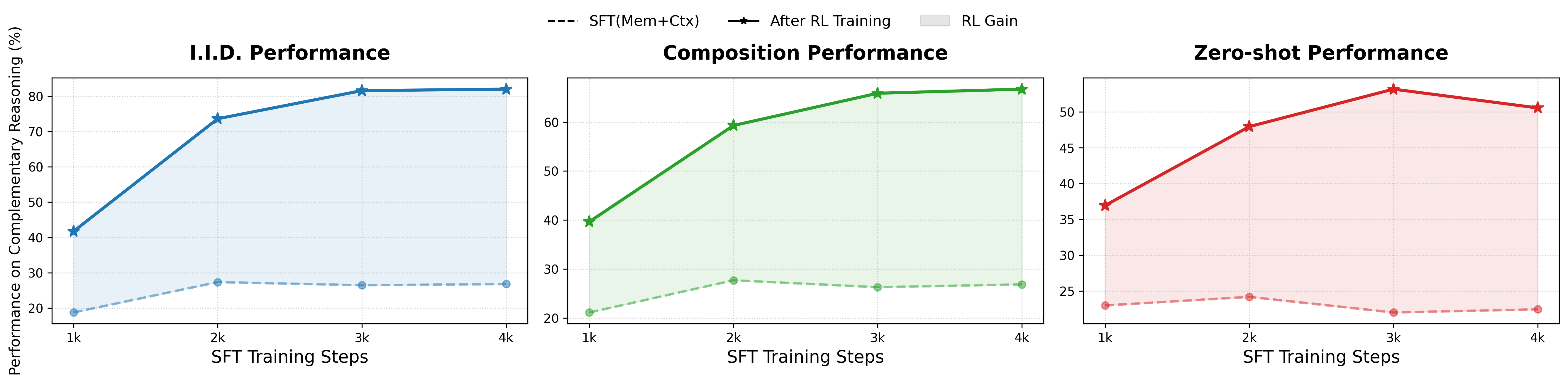}
    \caption{The impact of training steps (\textit{i.e.,} training loss) of the SFT model on RL generalization.}
    \label{fig:training_steps}
\end{figure*}

\section{Mechanistic Analysis}
\label{app:other_factors}

There are many factors affecting models' performance or generalizability. We showcase the sample efficiency and pass@$k$ performance as some features to see why the model with atomic skills are essentials for RL generalization in composite tasks. Here, we investigate from the perspective of training dynamics of \sftatomic, the impact of training loss of the base model for RL, the embedding distributions of the SFT and RL model and the uncertainty of the model.

\subsection{Training Dynamics of \sftatomic}
\label{app:training_dynamics}

\paragraph{Settings} To check when and how the ability of complementary reasoning ability (\comp) emerges through the training of parametric and contextual reasoning \textit{i.e.,} \sftatomic, we showcase the performance of \para, \context and \comp during the SFT training over \para and \context. We study Qwen-1.5B-Base and Qwen-0.5B-Base.

\paragraph{The ability of complementary reasoning emerges to some extent with the progress of both parametric and contextual reasoning.} Figure \ref{fig:training_dynamic} shows the training dynamic over training steps. First, it shows that as the SFT training with both \para and \context data progresses, the ability of \comp somehow emerges with \para and \context to some extent. Second, the conclusion is consistent over different model sizes, with larger models generalizes better to \comp with \para and \context data. Third, it further demonstrates that contextual reasoning is the easiest (learned from relatively early stage), while the parametric and complementary reasoning is relatively difficult.

\begin{figure*}[t]
    \centering
    \includegraphics[width=0.85\linewidth]{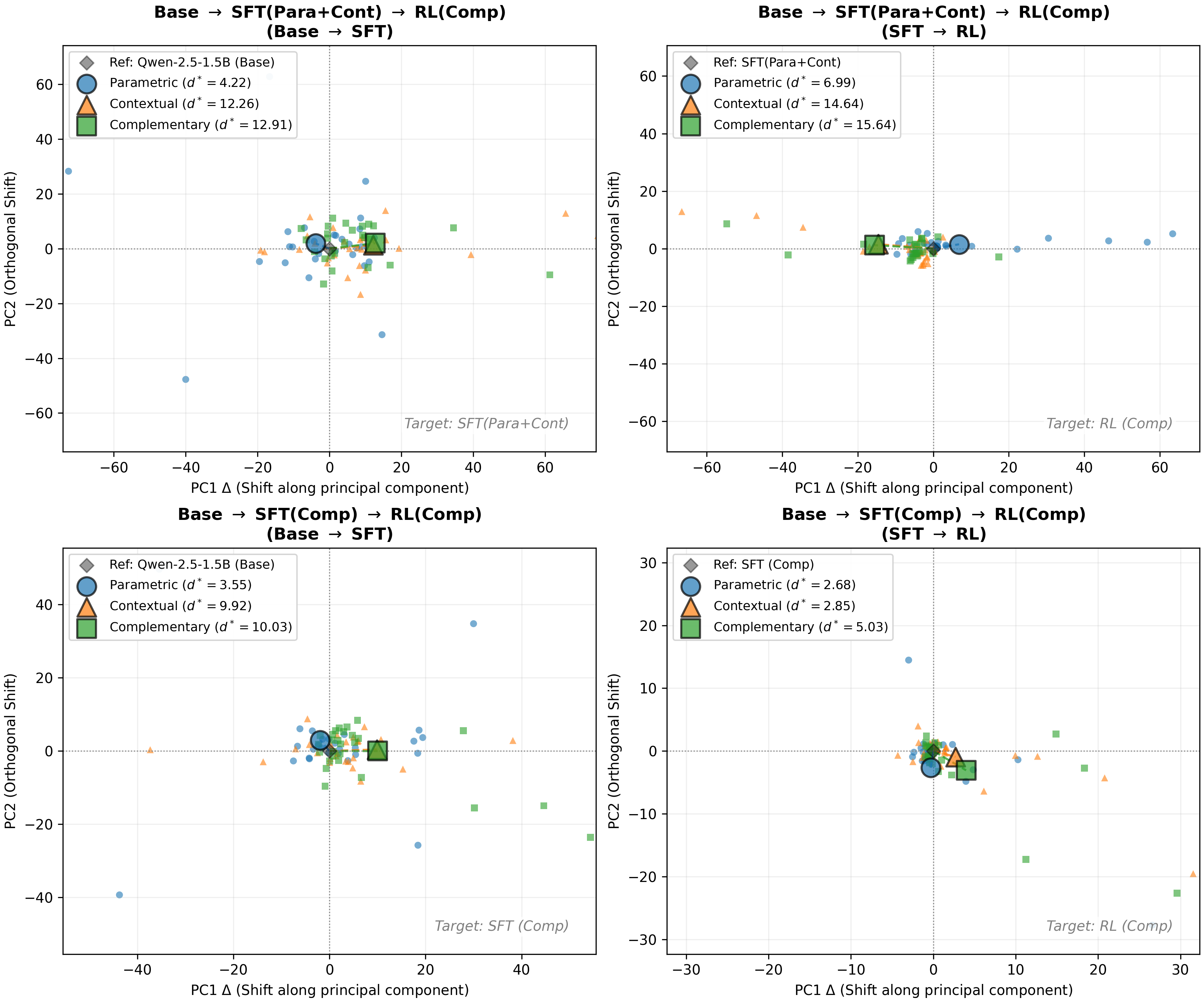}
    \caption{PCA Analysis of \sftatomic$\rightarrow$\rlcomp and \sftcomp$\rightarrow$\rlcomp. The scatter points represent the layer-wise shifts. The large markers represent the global centroid shift ($z^*$) for each reasoning type.}
    \label{fig:pca_shift}
\end{figure*}

\subsection{Effect of SFT Loss / Checkpoint}

We investigate a critical question regarding the interplay between Supervised Fine-Tuning (SFT) and Reinforcement Learning (RL): \textbf{Does the degree of SFT convergence dictate the model's potential for RL-based generalization?} Specifically, we aim to determine if a "grokking-like" phenomenon exists--where continuous optimization of SFT loss, even after apparent metric saturation, further unlocks the model's reasoning capabilities during the RL stage. Also, we study at which checkpoint (\textit{i.e.,} training loss) may the model emerge the ability of generalization. We take four intermediate checkpoints of \sftatomic to further conduct \rlcomp and evaluate the performance on three levels of generalization.

Figure~\ref{fig:training_steps} illustrates the performance trajectories across SFT checkpoints (1k to 4k steps), revealing three distinct phases of capability emergence:

$\bullet$ \textbf{Insufficient Representation.} 
    At the early stage (\textit{i.e.,} 1k steps, Training Loss $\approx 0.44$), the model has not yet internalized the necessary reasoning patterns. Consequently, it fails to generalize effectively during the RL stage, resulting in suboptimal performance across all metrics (e.g., Zero-shot accuracy is significantly lower compared to later stages).

$\bullet$ \textbf{Emergence of Capabilities.} 
    As the SFT loss decreases to approximately $0.16$ (\textit{i.e.,} 2k steps), we observe a sharp ``phase transition'' in downstream RL performance. While the base SFT model's direct performance (dashed lines) shows only moderate improvements, its \textit{latent potential} for RL adaptation increases dramatically. This suggests that the critical structures required for reasoning generalization are established during this interval.

$\bullet$ \textbf{Saturation and Robustness.} 
    Performance peaks around 3k steps (Loss $\approx 0.05$). Interestingly, further extending training to 4k steps—where the model nearly memorizes the training data (Loss $\approx 0.0004$)—does not yield further significant gains, nor does it lead to performance degradation. This indicates that while ``grokking'' (delayed generalization) effectively occurs between 1k and 3k steps, the benefit saturates once the loss drops below a certain threshold ($<0.05$). The model becomes robust, maintaining its high plasticity for RL even when deeply fitted to the SFT distribution.

In conclusion, \textbf{minimizing SFT loss is crucial up to a point}. The generalization capability for RL does not scale infinitely with lower loss but requires a sufficient "incubation" period (up to 3k steps in our setting) to fully emerge.

\begin{table*}[t]
    \centering
    \small
    \caption{Analysis of model uncertainty (measured by average entropy $\times 100$) across different training settings. \textbf{Overall} denotes average uncertainty on the test set, while the breakdown columns show uncertainty in IID, Compositional (Comp.), and Zero-shot settings. Lower values indicate higher model confidence.}
    \label{tab:uncertainty_analysis}
    \begin{tabular}{lcccc}
        \toprule
        \textbf{Setting} & \textbf{Overall} & \textbf{IID} & \textbf{Comp.} & \textbf{Zero-shot} \\
        \midrule
        \multicolumn{5}{l}{\textit{Effect of Training Steps (SFT on \para+\context)}} \\
        \sftatomic 3k step & 12.92 & 13.41 & 12.05 & 13.12 \\
        \sftatomic 4k step & 7.13 & 6.33 & 7.45 & 8.46 \\
        \midrule
        \multicolumn{5}{l}{\textit{10\% \comp data for \sftcomp, 90\% \comp data for \rlcomp}} \\
        \sftcomp& 8.93 & 6.56 & 9.89 & 12.96 \\
        \sftcomp$\rightarrow$\rlcomp & 8.99 & 8.49 & 8.97 & 10.17 \\
        {\sftatomic$\rightarrow$\rlcomp} & {2.58} & {1.63} & {2.54} & {4.84} \\
        \midrule
        \multicolumn{5}{l}{\textit{30\% \comp data for \sftcomp, 70\% \comp data for \rlcomp}} \\
        \sftcomp & 7.18 & 5.24 & 7.98 & 10.46 \\
        \sftcomp$\rightarrow$\rlcomp  & 6.11 & 4.85 & 6.36 & 8.65 \\
          \sftatomic$\rightarrow$\rlcomp & {2.90} & {1.96} & {2.84} & {5.17} \\
        \midrule
        \multicolumn{5}{l}{\textit{90\% \comp data for \sftcomp, 10\% \comp data for \rlcomp}} \\
       \sftcomp & 4.14 & 2.38 & 4.05 & 8.36 \\
        \sftcomp$\rightarrow$\rlcomp  & 4.11 & 3.27 & 3.97 & 6.25 \\
        \sftatomic$\rightarrow$\rlcomp & 4.10 & 3.23 & 3.88 & 6.46 \\
        \bottomrule
    \end{tabular}%
\end{table*}

\subsection{Latent Space PCA Analysis}

\paragraph{Setting: Layer-wise PCA Projection}
To investigate the internal mechanism behind the superior performance of \sftatomic$\rightarrow$\rlcomp compared to \sftcomp$\rightarrow$\rlcomp, we conduct a Principal Component Analysis (PCA) on the hidden states of the models. Our goal is to visualize how different training strategies affect the representation of Parametric, Contextual, and Complementary Reasoning.

Formally, for a given model pair (Anchor Model $M_{anc}$ and Target Model $M_{tgt}$) and a specific layer $l$, we extract the hidden states corresponding to the last token of the input queries. Let $\mathbf{H}_l^{anc} \in \mathbb{R}^{N \times D}$ and $\mathbf{H}_l^{tgt} \in \mathbb{R}^{N \times D}$ denote the hidden state matrices for $N$ queries at layer $l$, where $D$ is the hidden dimension.
To capture the relative shift induced by training, we fit the PCA transformation on $M_{anc}$'s states $\mathbf{H}_l^{anc}$, which defines a 2D coordinate system based on the principal variations of the reference model. We then project $M_{tgt}$'s states $\mathbf{H}_l^{tgt}$ into this fixed coordinate system. The shift vector for layer $l$ is calculated as the difference between the centroids of the projected target states and the anchor states. This process is repeated for all layers. Figure~\ref{fig:pca_shift} shows the layer-wise shifts by scatter points. The large markers represent the global centroid shift ($z^*$) for each Reasoning type.

\paragraph{Disentanglement of atomic skills in \sftatomic$\rightarrow$\rlcomp}
As shown in the top-left panel of Figure \ref{fig:pca_shift}, \sftatomic exhibits a significant ``disentanglement'' of atomic reasoning types. The centroid for \para data (Blue Circle) and \context data (Orange Triangle) move in distinct directions and magnitudes within the principal component space. This suggests that the \sftatomic stage effectively separates the internal representations required for parametric recall versus contextual reasoning.
Furthermore, in the subsequent RL stage (top-right panel), this separation is maintained and refined. Notably, the \comp data (Green Square) aligns closely with the established distributions of \para and \context data. This indicates that the model can effectively generalize the logic learned from \para and \context data to the composite \comp tasks.

\paragraph{Entanglement of skills in \sftcomp$\rightarrow$\rlcomp.}
In contrast, the bottom row reveals a phenomenon of ``representation entanglement''. Please note that the scale of the axis for the bottom row is lower than that for the top row. For the model trained only on \comp data, the embeddings for \para, \context, and \comp queries remain tightly clustered together, after both the SFT stage (bottom-left) and the RL stage (bottom-right). The centroids for all three data types are located close to the origin with overlapping distributions. This lack of separation implies that \sftcomp fails to distinguish between the underlying mechanisms of parametric and contextual reasoning, instead learning a coupled representation. 

We hypothesize that the superior generalization capability of \sftatomic$\rightarrow$\rlcomp stems from this structural disentanglement. By explicitly separating the latent representations of parametric and contextual capabilities during the SFT stage, the model establishes a robust basis that facilitates better adaptation during the RL phase. Conversely, the coupled representations in \sftcomp limit the model's ability to distinctly apply these capabilities, leading to suboptimal performance.

\subsection{Model Uncertainty}

We investigate the evolution of model uncertainty (quantified by the average prediction entropy $\times 100$) to understand the underlying dynamics of RL generalization. Table~\ref{tab:uncertainty_analysis} presents the entropy metrics across IID, Compositional, and Zero-shot subsets.

\begin{table*}[t]
    \centering
    \small
    \caption{Other empirical results on Complementary Reasoning test set.}
    \label{tab:other_sufficiency_results}
    \begin{tabular}{lccccc}
    \toprule
       \textbf{Setting}  &  \textbf{IID} & \textbf{Comp} &  \textbf{Zero-shot}  \\
       \midrule
       \sftatomic   & 35.18 & 28.20 & 24.07  \\
       \sftatomic$\rightarrow$\rlatomic   &  28.43&28.34&27.89 \\
       \sftatomic$\rightarrow$\rlatomic$\rightarrow$\rlcomp  & {74.00}&{62.47}&49.56 \\
      $\textsc{SFT}_{{\para+\context+\comp}}$ & \textbf{80.14}&\textbf{62.90}&43.25  \\

       \sftatomic$\rightarrow$\rlcomp  & 73.11&60.85&\textbf{50.87} \\
       \bottomrule
    \end{tabular}
\end{table*}

\textbf{Uncertainty is correlated with SFT convergence, not necessarily RL potential.}
We first address whether high uncertainty is a prerequisite for effective RL exploration. Comparing the \sftatomic checkpoints, the uncertainty drops significantly from 3k steps ($12.92$) to 4k steps ($7.13$) as the loss minimizes. Despite this lower starting entropy, we previously observed that the 4k-step model sustains high performance after RL. This indicates that a "calibrated" and confident SFT model (lower entropy) does not hinder subsequent RL generalization.
    
 \textbf{Task Difficulty Indicator.}
Consistently across all settings, the uncertainty is highest in the \textbf{Zero-shot} subset (e.g., $12.96$ for 10\% SFT(G3)). This aligns with intuition, as the model exhibits lower confidence on unseen distributions.
    
 \textbf{\sftatomic Facilitates Efficient RL Adaptation.}
    A key insight emerges when comparing the post-RL behaviors. In the 30\% data setting, \textit{30\% \sftcomp} ($7.18$) and \textit{\sftatomic} ($7.13$) start at remarkably similar uncertainty levels. However, after applying identical RL training, \sftcomp$\rightarrow$\rlcomp shows minimal entropy reduction ($7.18 \to 6.11$), suggesting the model struggles to find a more optimal, confident policy.
    In contrast, our recipe \sftatomic$\rightarrow$\rlcomp achieves a drastic reduction in uncertainty ($7.13 \to \mathbf{2.90}$).
This demonstrates that the model with sufficient atomic skills does not merely provide "randomness" for exploration; rather, it structures the latent space (see Figure \ref{fig:pca_shift}) in a way that allows RL to efficiently converge to a high-confidence, correct solution.

\section{Extended Sufficiency Results}
\label{app:sufficiency}

In \S \ref{sec:sufficiency}, we focus on whether our atomic-first recipe ``\sftatomic$\rightarrow$\rlcomp'' consistently yields superior generalization, compared to direct training on the target task ``\sftcomp$\rightarrow$\rlcomp'' which is the common practice in real-world complex tasks.
Here, we compare ``\textbf{\sftatomic$\rightarrow$\rlcomp}'' with other possible baselines: 1) ``\sftatomic'': SFT with the entire \para and \context data, which is the same as in Table \ref{tab:empirical_results_b}; 2) ``\textbf{\sftatomic$\rightarrow$\rlatomic}'': SFT with 80\% of \para and \context data and then RL with the rest 20\% of \para and \context data. The portion is based on our empirical results of splitting \comp data; 3) ``\textbf{\sftatomic$\rightarrow$\rlatomic$\rightarrow$\rlcomp}'': furthur RL with \comp based on ``\sftatomic$\rightarrow$\rlatomic''; 4) ``\textbf{$\textsc{SFT}_{\textsc{\para+\context+\comp}}$}'': SFT with the mix of \para, \context and \comp data. Specifically, for \rlcomp, we adopt the 12.8k random samples used in \S \ref{sec:necessity_training}. We show the performance on \comp test set in Table \ref{tab:other_sufficiency_results}.

It shows that \textbf{\sftatomic$\rightarrow$\rlatomic$\rightarrow$\rlcomp}  has very similar performance compared with our proposed \textbf{\sftatomic$\rightarrow$\rlcomp}, especially for Zero-shot setting. However, this does not contradict our conclusion, as the capture of sufficient atomic skills and RL training are essential for Zero-shot generalization and sufficient atomic skills are still the prerequisites. In addition, comparing \sftatomic with \textbf{\sftatomic$\rightarrow$\rlatomic}, we find that RL with atomic skills may not be beneficial to composite tasks. Note that for \para and \comp data, every test sample in \comp would be either Composition or Zero-shot level (these two baselines have never seen any \comp data during training. Moreover, Table \ref{tab:other_sufficiency_results} further enhances the sufficiency of our findings that our recipe \sftatomic$\rightarrow$\rlcomp is the key to generalization.

It is interesting that \textbf{$\textsc{SFT}_{\textsc{\para+\context+\comp}}$} also manages to generalize to Zero-shot scenarios to some extent and achieves comparable results with our \textbf{\sftatomic$\rightarrow$\rlcomp}. This shows that SFT can also generalize with careful mix of training data. However, this does not challenge our SFT Generalization Paradox that SFT directly with composite data fails to generalize \cite{wang2025generalization}. Furthermore, the setting itself is not as realistic-it is not trivial to obtain ample composite (\comp) data encompassing both \para and \context skills. Also, it is difficult to mix  \comp data to SFT training with all atomic skills. However, our \textbf{\sftatomic$\rightarrow$\rlcomp} assumes a common post-training practice with an initial SFT and a following RL stage. While the base model grasp the atomic skills (\textit{i.e.,} today's available LLMs), we focus on the training strategies and data mix when we have some composite data. And we demonstrate the condition of RL generalization.

\begin{table*}[t]
\centering
\caption{An example of human biography in the form of Python Dict and Natural Language Document.}
\label{tab:human_biography}
\small 
\renewcommand{\arraystretch}{1.3} 
\begin{tabular}{p{6cm}|p{5cm}} 
\toprule
\textbf{Python Dictionary} & \textbf{Natural Language Document} \\ 
\midrule
   \noindent "Allison Hill": \{ \\
    \hspace*{1em} "name": "Allison Hill", \\
    \hspace*{1em} "birth\_date": "1942-04-29", \\
    \hspace*{1em} "occupation": "Civil engineer, consulting", \\
    \hspace*{1em} "email": "garzaanthony@example.org", \\
    \hspace*{1em} "phone": "538.990.8386", \\
    \hspace*{1em} "new": true, \\
    \hspace*{1em} "died\_on": "2024-11-01", \\
    \hspace*{1em} "child": "Donald Marsh", \\
    \hspace*{1em} "pet": "Whiskers", \\
    \hspace*{1em} "wrote": "Baby administration", \\
    \hspace*{1em} "influenced\_by": "Matthew Cooper", \\
    \hspace*{1em} "mentoring": "Daniel Watkins", \\
    \hspace*{1em} "hobby": "painting", \\
    \hspace*{1em} "classmate": "Adam Villanueva", \\
    \hspace*{1em} "first\_language": "Finnish", \\
    \hspace*{1em} "roommate": "Shannon Krause", \\
    \hspace*{1em} "university": "University of Chicago", \\
    \hspace*{1em} "service": "Habitat for Humanity", \\
    \hspace*{1em} "known\_for": "painting", \\
    \hspace*{1em} "died\_in": "Brownbury", \\
    \hspace*{1em} "boss": "Lindsey Johnson", \\
    \hspace*{1em} "favorite\_food": "tacos" \\
    \} & \vspace{-10cm}  Allison Hill has a pet named Whiskers. Allison Hill spoke Finnish as their first language. A favorite activity of Allison Hill is painting. Lindsey Johnson is the boss of Allison Hill. Allison Hill was born on 1942-04-29. Allison Hill died on 2024-11-01. Allison Hill shared a room with Shannon Krause. Allison Hill penned Baby administration. Allison Hill was inspired by Matthew Cooper. The contact email for Allison Hill is garzaanthony@example.org. Allison Hill was famous for painting. Allison Hill was a member of Habitat for Humanity. Allison Hill mentors Daniel Watkins. Allison Hill's place of death was Brownbury. Allison Hill's phone number is 538.990.8386. Allison Hill works as a Civil engineer, consulting. Allison Hill is the parent of Donald Marsh. Allison Hill was a classmate of Adam Villanueva. Allison Hill loved eating tacos. Allison Hill went to University of Chicago.
    \\
\bottomrule
\end{tabular}
\end{table*}

\begin{table*}[]
\centering
\small
\caption{Templates for Relations in our Knowledge Graph.}
\label{tab:relation_templates}
\renewcommand{\arraystretch}{0.95} 
\begin{tabular}{l p{5cm} p{5.5cm}} 
\toprule
\textbf{Relation} & \textbf{Statement Template} & \textbf{Question Template} \\ 
\midrule
address & \{e1\} resides at \{e2\}. & What is \{e1\}'s address? \\
awards & \{e1\} won the \{e2\} award. & What awards has \{e1\} won? \\
best\_friend & \{e1\}'s best friend is \{e2\}. & Who is \{e1\}'s closest friend? \\
birth\_date & \{e1\} was born on \{e2\}. & When was \{e1\} born? \\
birth\_place & \{e1\} hails from \{e2\}. & Where was \{e1\} born? \\
boss & \{e1\} works under \{e2\}. & Who is \{e1\}'s boss? \\
boss\_of & \{e1\} manages \{e2\}. & Who works under \{e1\}? \\
child & \{e2\} is the child of \{e1\}. & Who is \{e1\}'s child? \\
classmate & \{e1\} studied alongside \{e2\}. & Who attended school with \{e1\}? \\
colleague & \{e1\} works alongside \{e2\}. & Who are \{e1\}'s colleagues? \\
died\_in & \{e1\} passed away in \{e2\}. & Where did \{e1\} die? \\
died\_on & \{e1\} died on \{e2\}. & When did \{e1\} pass away? \\
email & You can reach \{e1\} at \{e2\}. & What is \{e1\}'s email address? \\
favorite\_food & \{e1\}'s favorite dish was \{e2\}. & What food does \{e1\} enjoy the most? \\
first\_language & \{e1\}'s native language was \{e2\}. & What is \{e1\}'s first language? \\
hobby & A favorite activity of \{e1\} is \{e2\}. & What does \{e1\} enjoy doing? \\
influence & \{e1\} shaped the career of \{e2\}. & Who was influenced by \{e1\}? \\
influenced\_by & \{e1\} looked up to \{e2\}. & Who inspired \{e1\}? \\
known\_for & \{e1\} gained recognition for \{e2\}. & What is \{e1\} famous for? \\
leader\_of & \{e1\} was the leader of \{e2\}. & Which group was \{e1\} in charge of? \\
lived\_in & \{e1\} resided in \{e2\}. & Where has \{e1\} lived? \\
major & \{e1\} majored in \{e2\}. & What did \{e1\} specialize in? \\
mentored\_by & \{e1\} received guidance from \{e2\}. & Who mentored \{e1\}? \\
mentoring & \{e2\} is a student of \{e1\}. & Who does \{e1\} mentor? \\
nationality & \{e1\} is a citizen of \{e2\}. & What is \{e1\}'s nationality? \\
neighbor & \{e1\} lives next to \{e2\}. & Who resides beside \{e1\}? \\
occupation & \{e1\} is employed as \{e2\}. & What does \{e1\} do for a living? \\
parent & \{e1\}'s parent is \{e2\}. & Who is the parent of \{e1\}? \\
pet & \{e1\} owns a pet called \{e2\}. & What is the name of \{e1\}'s pet? \\
philanthropy & \{e1\} donated to \{e2\}. & Which causes did \{e1\} support? \\
phone & \{e1\} can be reached at \{e2\}. & What is \{e1\}'s phone number? \\
rival & \{e1\} had a rivalry with \{e2\}. & Who did \{e1\} compete with? \\
roommate & \{e1\} shared a room with \{e2\}. & Who lived with \{e1\}? \\
service & \{e1\} was a member of \{e2\}. & Which organization did \{e1\} serve in? \\
sibling & \{e1\} and \{e2\} are siblings. & Who are \{e1\}'s siblings? \\
spouse & \{e1\} is married to \{e2\}. & Who is \{e1\}'s spouse? \\
university & \{e1\} went to \{e2\}. & Which university did \{e1\} attend? \\
worked\_at & \{e1\} held a position at \{e2\}. & Where did \{e1\} work? \\
wrote & \{e1\} authored the book \{e2\}. & Which book did \{e1\} write? \\
\bottomrule
\end{tabular}
\end{table*}

\section{Terminologies}

See Table \ref{tab:terminologies}.
\begin{table*}[h]
    \centering
    \small
    \caption{Glossary of Symbols and Abbreviations}
    \label{tab:symbols_glossary}
    \setlength{\tabcolsep}{10pt} 
    \begin{tabular}{p{0.27\textwidth} p{0.70\textwidth}}
        \toprule
        \textbf{Symbol/Abbreviation} & \textbf{Description} \\
        \midrule
        
        \para & Parametric reasoning task. \\
        \context & Contextual reasoning task. \\
        \comp & Complementary reasoning task. \\
        $\mathcal{C}_{\text{\para}}$ & Parametric reasoning skill. \\
        $\mathcal{C}_{\text{\context}}$ & Contextual reasoning skill. \\
        $\mathcal{C}_{\text{\comp}}$ & Complementary reasoning skill. \\
        RL & Reinforcement Learning, one of the core training strategies discussed in this work. \\
        SFT & Supervised Fine-Tuning, one of the core training strategies discussed in this work. \\
        LoRA & Low-Rank Adaptation, one of the training strategies discussed in this work. \\
        
        \sftatomic & The model obtained after performing SFT on the combined training set of \para and \context. \\
        \sftcomp & The model obtained after performing SFT on the training set of \comp. \\
        
        $\textsc{SFT}_{10\%\comp}$ & The model obtained after performing SFT on 10\% random samples of \comp training set. \\

        \sftatomic$\rightarrow$\rlcomp & The model obtained after performing RL on the training set of \comp from \sftatomic. \\
        \sftcomp$\rightarrow$\rlcomp & The model obtained after performing RL on the training set of \comp from \sftcomp. \\
        IID & Independent and Identically Distributed generalization type in the test set. \\

        
        \bottomrule
    \end{tabular}
    \label{tab:terminologies}
\end{table*}

\end{document}